\documentclass{article}

\usepackage{microtype}
\usepackage{graphicx}
\usepackage{subfigure}
\usepackage{booktabs} 

\usepackage{hyperref}

\usepackage[accepted]{icml2018}

\icmltitlerunning{Improving Optimization for Models With Continuous Symmetry Breaking}

\usepackage{color}
\definecolor{royalblue}{rgb}{0.0, 0.14, 0.4}
\hypersetup{citecolor=royalblue}

\usepackage{amsmath}
\usepackage{amsfonts}
\usepackage{tabularx}
\usepackage[algoruled,algo2e]{algorithm2e}
\usepackage[percent]{overpic}
\newcommand\Ell{\mathcal L}

\newcommand\bX{\mathbf X}
\newcommand\bZ{\mathbf Z}

\newcommand\bL{\mathbf L}

\newcommand\bR{\mathbf R}

\newcommand\bg{\boldsymbol\Gamma}

\newcommand\IR{\mathbb R}

\begin{document}

\twocolumn[
\icmltitle{Improving Optimization for Models With Continuous Symmetry Breaking}



\icmlsetsymbol{equal}{*}

\begin{icmlauthorlist}
\icmlauthor{Robert Bamler}{dr}
\icmlauthor{Stephan Mandt}{dr}
\end{icmlauthorlist}

\icmlaffiliation{dr}{Disney Research, Glendale, CA, USA}

\icmlcorrespondingauthor{Robert Bamler}{robert.bamler@gmail.com}
\icmlcorrespondingauthor{Stephan Mandt}{stephan.mandt@gmail.com}

\icmlkeywords{Machine Learning, ICML, Optimization, Symmetries, Goldstone}

\vskip 0.3in
]


\printAffiliationsAndNotice{}  


\begin{abstract}
Many loss functions in representation learning are invariant under a continuous symmetry transformation.
For example, the loss function of word embeddings~\citep{mikolov_distributed_2013} remains unchanged if we simultaneously rotate all word and context embedding vectors.
We show that representation learning models for time series possess an approximate continuous symmetry that leads to slow convergence of gradient descent.
We propose a new optimization algorithm that speeds up convergence using ideas from gauge theory in physics.
Our algorithm leads to orders of magnitude faster convergence and to more interpretable representations, as we show for dynamic extensions of matrix factorization and word embedding models.
We further present an example application of our proposed algorithm that translates modern words into their historic equivalents.
\end{abstract}

\section{Introduction}

Symmetries frequently occur in machine learning.
They express that the loss function of a model is invariant under a certain group of transformations.
For example, the loss function of matrix factorization or word embedding models remains unchanged if we simultaneously rotate all embedding vectors with the same rotation matrix.
This is an example of a \emph{continuous} symmetry, since the rotations are parameterized by a continuum of real-valued angles.

Sometimes, the symmetry of a loss function is broken, e.g., due to the presence of an additional term that violates the symmetry.
For example, this may be a weak regularizer.
In this paper, we show that such symmetry breaking may induce slow convergence problems in gradient descent, in particular when the symmetry breaking is weak.
We solve this problem with a new optimization algorithm.

Weak continuous symmetry breaking leads to an ill-conditioned optimization problem.
When a loss function is invariant under a continuous symmetry, it has a manifold of \emph{degenerate} (equivalent) minima.
This is usually not a problem because any such minimum is a valid solution of the optimization problem.
However, adding a small symmetry breaking term to the loss function lifts the degeneracy and forces the model to prefer one minimum over all others.
As we show, this leads to an ill-conditioned Hessian of the loss, with a small curvature along symmetry directions and a large curvature perpendicular to them.
The ill-conditioned Hessian results in slow convergence of gradient descent.

We propose an optimization algorithm that speeds up convergence by separating the optimization in the symmetry directions from the optimization in the remaining directions.
At regular intervals, the algorithm efficiently minimizes the small symmetry breaking term in such a way that the minimization does not degrade the symmetry invariant term.

Symmetries can be broken explicitly, e.g., due to an additional term such as an $L_1$ regularizer in a word embedding model~\citep{sun2016sparse}. However, perhaps more interestingly, symmetries can also be broken by couplings between model parameters.
This is known as \emph{spontaneous} symmetry breaking in the physics community.

One of our main findings is that spontaneous symmetry breaking occurs in certain time series models, such as dynamic matrix factorizations and dynamic word embedding models~\citep{lu2009aspatiot,koren2010collaborative,charlin2015dynamic,bamler_dynamic_2017,rudolph2017dynamic}.
In these models, it turns out that model parameters may be smoothly twisted along the time axis, and that these twists contribute only little to the loss, thus leading to a small gradient.
These inexpensive smooth twists are known in the physics community as Goldstone modes \citep{altland2010condensed}.

Our contributions are as follows:
\begin{itemize}
  \item We identify a broad class of models that suffer from slow convergence of gradient descent due to Goldstone modes.
  We explain both mathematically and pictorially how Goldstone modes lead to slow convergence.
  \item Using ideas from gauge theories in physics, we propose Goldstone Gradient Descent (Goldstone-GD), an optimization algorithm that speeds up convergence by separating the optimization along symmetry directions from the remaining coordinate directions.
  \item We evaluate the Goldstone-GD algorithm experimentally with dynamic matrix factorizations and Dynamic Word Embeddings.
  We find that Goldstone-GD converges orders of magnitude faster and finds more interpretable embedding vectors than standard gradient descent (GD) or GD with diagonal preconditioning.
  \item For Dynamic Word Embeddings~\citep{bamler_dynamic_2017}, Goldstone-GD allows us to find historic synonyms of modern English words, such as ``wagon'' for ``car''.
  Without our advanced optimization algorithm, we were not able to perform this task.
\end{itemize}

Our paper is structured as follows.
Section~\ref{sec:related} describes related work.
In Section~\ref{sec:problem-setting}, we specify the model class under consideration, introduce concrete example models, and discuss the slow convergence problem.
In Section~\ref{sec:method}, we propose the Goldstone-GD algorithm that solves the slow convergence problem.
We report experimental results in Section~\ref{sec:experiments} and provide concluding remarks in Section~\ref{sec:conclusions}.


\section{Related Work}
\label{sec:related}

Our paper discusses continuous symmetries in machine learning and proposes a new optimization algorithm.
In this section, we summarize related work on both aspects.

Most work on symmetries in machine learning focuses on discrete symmetries.
Convolutional neural networks \citep{Lecun98gradient} exploit the discrete translational symmetry of images.
This idea was generalized to arbitrary discrete symmetries \citep{gens2014deep}, to the permutation symmetry of sets \citep{zaheer2017deep}, and to discrete symmetries in graphical models \citep{bui2012automorphism,noessner2013rockit}.
Discrete symmetries do not cause an ill-conditioned optimization problem because they lead to isolated degenerate minima rather than a manifold of degenerate minima.

In this work, we consider models with continuous symmetries.
Continuous rotational symmetries have been identified in deep neural networks~\citep{badrinarayanan2015understanding}, matrix factorization~\citep{mnih2008probabilistic,gopalan2015scalable}, linear factor models~\citep{murphy2012machine}, and word embeddings \citep{mikolov_efficient_2013,mikolov_distributed_2013,pennington2014glove,barkan2016bayesian}.
Dynamic matrix factorizations \citep{lu2009aspatiot,koren2010collaborative,sun2012dynamic,charlin2015dynamic} and dynamic word embeddings \citep{bamler_dynamic_2017,rudolph2017dynamic} generalize these models to sequential data.
These are the models whose optimization we address in this paper.
A specialized optimization algorithm for a loss function with a continuous symmetry was presented in~\citep{choi1999natural}.
Our discussion is more general since we only require invariance under a collective rotation of all feature vectors, and not under independent symmetry transformations of each individual feature.

The slow convergence in these models is caused by shallow directions of the loss function.
Popular methods to escape a shallow valley of a loss function~\citep{duchi_adaptive_2011,zeiler2012adadelta,kingma_adam_2014} use diagonal preconditioning.
As confirmed by our experiments, diagonal preconditioning does not speed up convergence when the shallow directions correspond to collective rotations of many model parameters, which are not aligned with the coordinate axes.

Natural gradients \citep{amari1998natural,martens2014new} are a more sophisticated form of preconditioning, which has been applied to deep learning \citep{pascanu2013revisiting} and to variational inference \citep{hoffman_stochastic_2013}.
Our proposed algorithm uses natural gradients in a subspace where they are cheap to obtain.
Different to the Krylov subspace method \citep{vinyals2012krylov}, we construct the subspace such that it always contains the shallow directions of the loss.


\section{Problem Setting}
\label{sec:problem-setting}

In this section, we formalize the notion of continuous symmetry breaking (Section~\ref{sec:weaksymbr}), and we specify the type of models that we investigate in this paper (Section~\ref{sec:models}).
We then show that the introduced models exhibit a specific type of symmetry breaking that is generically weak, which leads to slow convergence of gradient descent (Section~\ref{sec:spontsymbr}).

\subsection{Symmetry Breaking in Representation Learning}
\label{sec:weaksymbr}

\begin{figure*}[t]
\vskip 0.04in
\begin{center}
\centerline{\begin{overpic}[width=\textwidth]{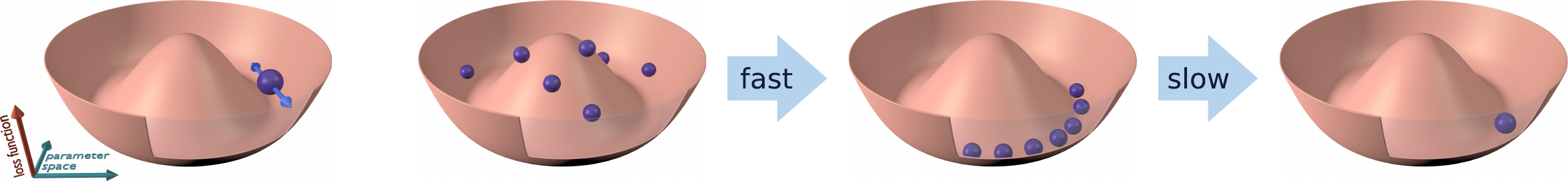}
 \put (0,9.9) {\small a)\hspace{39.4mm} b)\hspace{43.7mm} c)\hspace{43.7mm} d)}
\end{overpic}}
\caption{
a) A rotationally symmetric loss function $\ell$ has a manifold of degenerate minima, and zero curvature tangential to this manifold.
b-d) Gradient descent converges in two phases.
b) random initialization;
c) Goldstone mode;
d) minimum of the total loss function $\Ell$.}
\label{fig:champagne_bottle}
\end{center}
\vskip -0.2in
\end{figure*}


We formalize the notion of weak continuous symmetry breaking
and show that it leads to an ill-conditioned optimization problem.
The red surface in Figure~\ref{fig:champagne_bottle}a illustrates a rotationally symmetric loss~$\ell$.
It has a ring of degenerate (i.e., equivalent) minima.
The purple sphere depicts one arbitrarily chosen minimum.
Tangential to the ring of degenerate minima, i.e., along the blue arrows, $\ell$ is flat. 

For a machine learning example, consider factorizing a large matrix $X$ into the product $U^\top V$ of two smaller matrices $U$ and $V$ by minimizing the loss $\ell(U,V)=||X-U^\top V||_2^2$.
Rotating all columns of $U$ and $V$ by the same orthogonal%
\footnote{We call a square matrix $R$ `orthogonal' if $R^\top\! R$ is the identity. This is sometimes also called an `orthonormal' matrix.}
rotation matrix $R$ such that $U \gets RU$ and $V \gets RV$ does not change $\ell$ since $(RU)^{\!\top}\! RV = U^\top \! (R^\top \! R)V = U^\top V$.

The continuous rotational symmetry of $\ell$ leads to a manifold of degenerate minima:
if $(U^*, V^*)$ minimizes~$\ell$, then so does $(RU^*, RV^*)$ for any rotation matrix $R$.
On the manifold of degenerate minima, the gradient assumes the constant value of zero.
A constant gradient (first derivative) means that the curvature (second derivative) is zero.
More precisely, the Hessian of~$\ell$ has a zero eigenvalue for all eigenvectors that are tangential to the manifold of degenerate minima.
Usually, a zero eigenvalue of the Hessian indicates a maximally ill-conditioned optimization problem, but this is not an issue here.
The zero eigenvalue only means that convergence within the manifold of degenerate minima is infinitely slow.
This is of no concern since any minimum is a valid solution of the optimization problem.

A problem arises when the continuous symmetry is weakly broken, e.g., by adding a small $L_1$ regularizer to the loss.
A sufficiently small regularizer changes the eigenvalues of the Hessian only slightly, leaving it still ill-conditioned.
However, even a small regularizer lifts the degeneracy and turns the manifold of exactly degenerate minima into a shallow valley with one preferred minimum.
Convergence along this shallow valley is slow because of the ill-conditioned Hessian.
This is the slow convergence problem that we address in this paper.
We present a more natural setup that exhibits this problem in Sections~\ref{sec:models}-\ref{sec:spontsymbr}.
Our solution, presented in Section~\ref{sec:method}, is to separate the optimization in the symmetry directions from the optimization in the remaining directions.

\subsection{Representation Learning for Time Series}
\label{sec:models}

We define a broad class of representation learning models for sequential data, and introduce three example models that are investigated experimentally in this paper.
As we show in Section~\ref{sec:spontsymbr}, the models presented here suffer from slow convergence due to a specific kind of symmetry breaking.

\paragraph{General Model Class.}
We consider data $\bX \equiv \{X_t\}_{t=1:T}$ that are associated with additional metadata $t$, such as a time stamp.
For each $t$, the task is to learn a low dimensional representation $Z_t$ by minimizing what we call a `local loss function' $\ell(X_t; Z_t)$.
We add a quadratic regularizer $\psi(\bZ)$ that couples the representations $\bZ\equiv\{Z_t\}_{t=1:T}$ along the \hbox{$t$-dimension}.
In a Bayesian setup, $\psi$ comes from the log-prior of the model.
The total loss function is thus
\begin{align} \label{eq:multitask-loss-1}
    \Ell(\bZ) &= \sum_{t=1}^T \ell(X_t; Z_t) + \psi(\bZ).
\end{align}
For each task $t$, the representation $Z_t$ is a matrix whose columns are embedding vectors of some low dimension $d$.
We assume that $\ell$ is invariant under a collective rotation of all columns of $Z_t$:
let $R$ be an arbitrary orthogonal rotation matrix of the same dimension as the embedding space, then
\begin{align}
    \label{eq:symmetry}
    \ell(X_t; R Z_t) = \ell(X_t; Z_t).
\end{align}
Finally, we consider a specific form of the regularizer $\psi$ which is quadratic in $\bZ$, and which is defined in terms of a sparse symmetric coupling matrix $\bL \in \mathbb{R}^{T\times T}$:
\begin{align}
\label{eq:prior_abstract}
    \psi(\bZ) = \tfrac12 {\rm Tr} (\bZ^\top \bL \bZ).
\end{align}
Here, the matrix-vector multiplications are carried out in $t$-space, and the trace runs over the remaining dimensions.
Note that, different to Section~\ref{sec:weaksymbr}, we do not require $\psi$ to have a small coefficient.
We only require the coupling matrix $\bL$ to be sparse.
In the examples below, $\bL$ is tridiagonal and results from a Gaussian Markovian time series prior.
In a more general setup, $\bL = \mathbf D-\mathbf A$ is the Laplacian matrix of a sparse weighted graph~\citep{poignard2018spectra}.
Here, $\mathbf A$ is the adjacency matrix, whose entries are the coupling strengths, and the degree matrix $\mathbf D$ is diagonal and defined such that the entries of each row of $\bL$ sum up to zero.

Equations~\ref{eq:multitask-loss-1}, \ref{eq:symmetry}, and \ref{eq:prior_abstract} specify the problem class of interest in this paper.
The following paragraphs introduce the specific example models used in our experiments.
In Section~\ref{sec:spontsymbr}, we show that the sparse coupling in these models leads to weak continuous symmetry breaking and therefore to slow convergence of gradient descent (GD).

\paragraph{Model 1: Dense Dynamic Matrix Factorization.}
Consider the task of factorizing a large matrix $X_t$ into a product $U_t^\top V_t$ of two smaller matrices.
The latent representation is the concatenation of the two embedding matrices,
\begin{align} \label{eq:matfact-z}
  Z_t\equiv(U_t, V_t).
\end{align}
In a Gaussian matrix factorization, the local loss function is
\begin{align} \label{eq:dmf-loss}
    \ell(X_t; Z_t) &= - \log \mathcal N(X_t; U_t^\top V_t, I)
\end{align}
In dynamic matrix factorization models, the data $\bX$ are observed sequentially at discrete time steps $t$, and the representations $\bZ$ capture the temporal evolution of latent embedding vectors.
We use a Markovian Gaussian time series prior with a coupling strength $\lambda$, resulting in the regularizer
\begin{align}
    \label{eq:time-series-prior}
    \psi(\bZ) &= \frac{\lambda}{2} \sum_{i=1}^N
        \sum_{t=1}^{T-1} ||z_{t+1,i} - z_{t,i}||_2^2.
\end{align}
Here, the vector $z_{t,i}$ is the $i$\textsuperscript{th} column of the matrix $Z_t$, i.e., the $i$\textsuperscript{th} embedding vector, and $N$ is the number of columns.
The regularizer allows the model to share statistical strength across time.
By multiplying out the square, we find that $\psi$ has the form of Eq.~\ref{eq:prior_abstract} with a tridiagonal coupling matrix,
\begin{align} \label{eq:tridiagonal-l}
    \bL &= \lambda \begin{pmatrix}
        1 & -1 & & & \\
        -1 & 2 & -1 & & \\
        & \ddots & \ddots & \ddots & \\
        & & -1 & 2 & -1 \\
        & & & -1 & 1
    \end{pmatrix}.
\end{align}

\paragraph{Model 2: Sparse Dynamic Matrix Factorization.}
In a sparse matrix factorization, the local loss $\ell$ involves only few components of the matrix $U_t^\top V_t$.
The latent representation is again $Z_t\equiv(U_t,V_t)$.
We consider a model for movie ratings where each user rates only few movies.
When user~$i$ rates movie $j$ in time step $t$, we model the log-likelihood to obtain the binary rating $x\in\{\pm1\}$ with a logistic regression,
\begin{align} \label{eq:log-reg}
  \log p(x|u_{t,i}, v_{t,j}) = \log \sigma(x\, u_{t,i}^\top v_{t,j})
\end{align}
with the sigmoid function $\sigma(\xi) = 1/(1+e^{-\xi})$.
Eq.~\ref{eq:log-reg} is the log-likelihood of the rating $x$ of a single movie by a single user.
We obtain the full log-likelihood $\log p(X_t | Z_t)$ for time step $t$ by summing over the log-likelihoods of all ratings observed at time step $t$.
The local loss is
\begin{align} \label{eq:sparse-matfact-localloss}
  \ell(X_t; Z_t) = -\log p(X_t|Z_t) + \frac{\gamma}{2} {||Z_t||}_2^2.
\end{align}
Here, ${||\cdot||}_2$ is the Frobenius norm, and we add a quadratic regularizer with strength $\gamma$ since data for some users or movies may be scarce.
We distinguish this local regularizer from the time series regularizer $\psi$, given again in Eq.~\ref{eq:time-series-prior}, as the local regularizer does not break the rotational symmetry.

\paragraph{Model 3: Dynamic Word Embeddings.}
Word embeddings map words from a large vocabulary to a low dimensional representation space such that neighboring words are semantically similar, and differences between word embedding vectors capture syntactic and semantic relations.
We consider the Dynamic Word Embeddings model \citep{bamler_dynamic_2017}, which uses a probabilistic interpretation of the Skip-Gram model with negative sampling, also known as word2vec \citep{mikolov_distributed_2013,barkan2016bayesian}, and combines it with a time series prior.
The model is trained on $T$ text sources with time stamps $t$, and it assigns two time dependent embedding vectors $u_{t,i}$ and $v_{t,i}$ to each word $i$ from a fixed vocabulary.
The embedding vectors are obtained by simultaneously factorizing two matrices, which contain so-called positive and negative counts of word-context pairs.
Therefore, the representation $Z_t\equiv(U_t,V_t)$ for each time step is invariant under orthogonal transformations.
The regularizer $\psi$ comes from the time series prior, which is a discretized Ornstein-Uhlenbeck process, i.e., it combines a random diffusion process with a local quadratic regularizer.

\subsection{Symmetry Breaking in Representation Learning for Time Series}
\label{sec:spontsymbr}

We show that the time series models introduced in Section~\ref{sec:models} exhibit a variant of the symmetry breaking discussed in Section~\ref{sec:weaksymbr}.
This variant of symmetry breaking is generically weak, thus causing slow convergence.
We discuss the convergence first geometrically and then more formally.

\paragraph{Geometric Picture: Goldstone Modes.}
We find that minimizing the total loss $\Ell$ in Eq.~\ref{eq:multitask-loss-1} with GD converges in two phases, illustrated in Figures~\ref{fig:champagne_bottle}b-d.
Each purple sphere depicts an embedding vector $Z_t$ for a time step~$t$ of the model.
The red surface is the rotationally symmetric local loss function~$\ell$.
For a simpler visualization, we assume here that $\ell$ is the same function for every time step~$t$, and that each $Z_t$ contains only a single embedding vector.


In the first phase, GD starts from a random initialization (Figure~\ref{fig:champagne_bottle}b) and quickly finds approximate minima of the local loss functions $\ell$ for each time step.
At this stage, we observe in experiments that the parameters of the time series model twist smoothly around the rotation center of~$\ell$ (Figure~\ref{fig:champagne_bottle}c).
Such a configuration is called a Goldstone mode in the physics literature \citep{altland2010condensed}.

The coupling~$\psi$ in Eqs.~\ref{eq:multitask-loss-1} and~\ref{eq:time-series-prior} penalizes Goldstone modes as it tries to pull neighboring spheres together.
In the second phase of convergence, GD eliminates Goldstone modes and arrives eventually at the minimum of the total loss $\Ell$ (Figure~\ref{fig:champagne_bottle}d; in this toy example, the chain contracts to a single point).
Convergence in this phase is slow because, as we show below, the gradient of $\Ell$ is suppressed in a Goldstone mode (Goldstone theorem) \citep{altland2010condensed}.

\begin{figure}[t]
\vskip 0.04in
\begin{center}
\centerline{\includegraphics[width=\columnwidth]{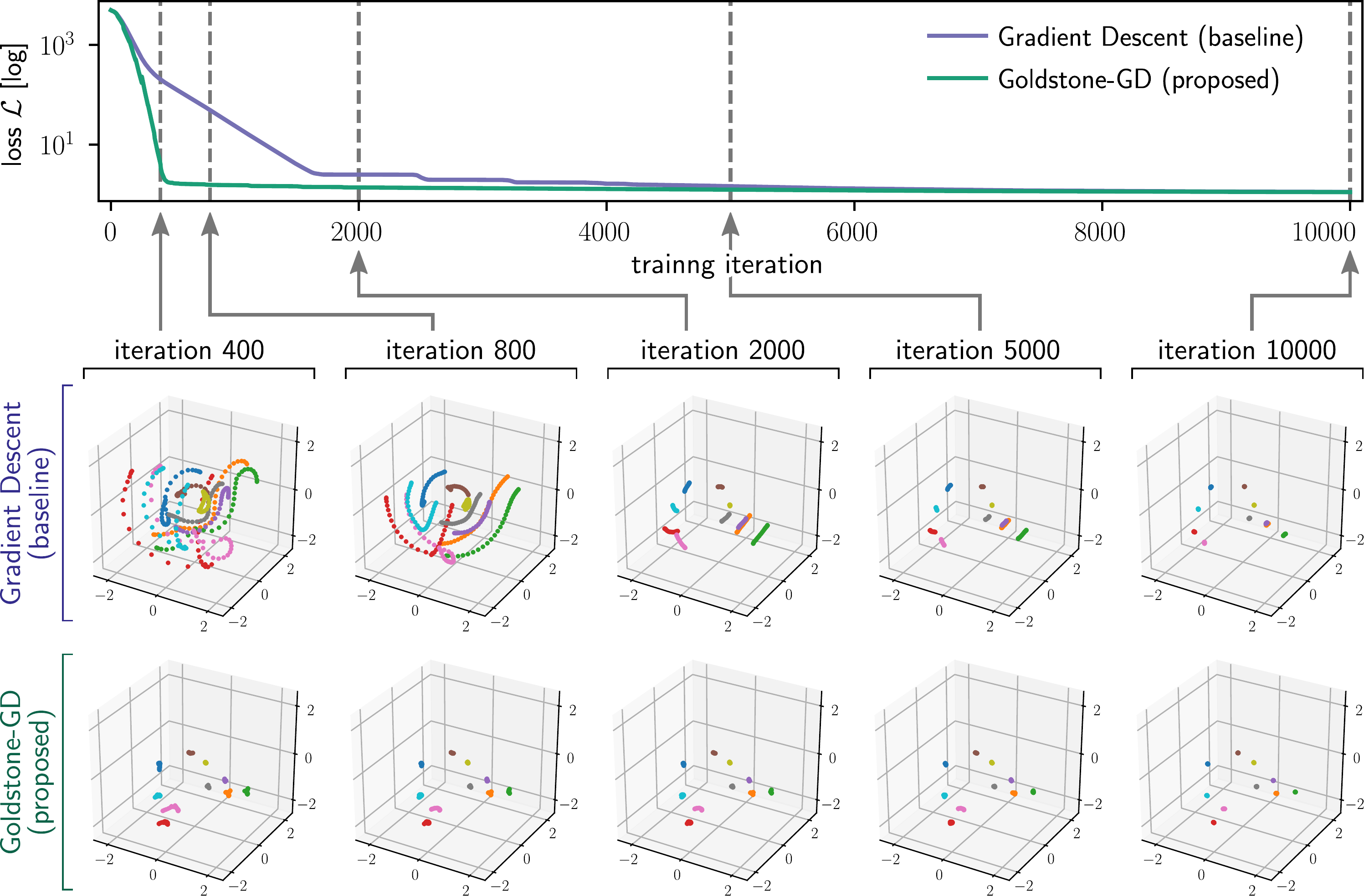}}
\caption{Goldstone modes and slow convergence of GD in a dynamic matrix factorization (see Section~\ref{sec:exp-artificial-data}).
Colored points in $3d$ plots show each embedding vector for all time steps of the model.}
\label{fig:matrixfact-mini}
\end{center}
\vskip -0.2in
\end{figure}

Figure~\ref{fig:matrixfact-mini} identifies the same two phases of convergence in an experiment.
The $3d$ plots show snapshots of the embedding space in a Gaussian dynamic matrix factorization with a $3d$ representation space and $T=30$ time steps (details in Section~\ref{sec:exp-artificial-data}).
Points of equal color show the trajectory of one embedding vector, i.e., $z_{t,i}$ for $t\in\{1,\ldots,T\}$ and fixed~$i$.

We see that GD (top row of $3d$ plots) arrives at a Goldstone mode, i.e., smooth twists around the origin, after about $400$ training iterations.
In this toy experiment, the local loss~$\ell$ is again identical for all~$t$.
Thus, in the optimum, each trajectory contracts to a single point, but this contraction takes many more training iterations.
By contrast, our algorithm evades Goldstone modes and converges much faster.

\paragraph{Formal Picture: Eigenvalues of the Hessian.}
Goldstone modes decay slowly in GD because the gradient of the total loss $\Ell$ is suppressed in a Goldstone mode.
This can be seen by analyzing the eigenvalues of the Hessian $\mathbf H$ of $\Ell$ at its minimum.
For a configuration $\bZ$ that is close to the true minimum $\bZ^*$, the gradient is approximately $\mathbf H (\bZ - \bZ^*)$.

The Hessian of $\Ell$ in Eq.~\ref{eq:multitask-loss-1} is the sum of the Hessians of the local loss functions $\ell$ plus the Hessian of the regularizer~$\psi$.
As discussed in Section~\ref{sec:weaksymbr}, the Hessians of $\ell$ all have exact zero eigenvalues along the symmetry directions.
Within this nullspace, only the Hessian $\mathbf H^{(\psi)}$ of the regularizer $\psi$ remains.
From Eq.~\ref{eq:prior_abstract}, we find $\mathbf H^{(\psi)} = \bL \otimes I_{N\times N} \otimes I_{d\times d}$ where $\otimes$ is the tensor product, and $I_{N\times N}$ and $I_{d\times d}$ are identity matrices in the input and embedding space, respectively.
Thus, $\mathbf H^{(\psi)}$ has the same eigenvalues as the Laplacian matrix~$\bL$ of the coupling graph, each with multiplicity $Nd$.

Since the rows of $\bL$ sum up to zero, $\bL$ has a zero eigenvalue for the eigenvector $(1, \ldots, 1)^\top$.
This is because the total loss $\Ell$ is exactly invariant under global rotations of all embeddings~$\bZ$ by the same rotation matrix.
As discussed in Section~\ref{sec:weaksymbr}, zero eigenvalues due to an \emph{exact} continuous symmetry do not induce slow convergence of GD.

The speed of convergence is governed by the lowest nonzero eigenvalue of the Hessian, and therefore of~$\bL$.
In a Markovian time series model, $\bL$ in Eq.~\ref{eq:tridiagonal-l} couples neighbors along a chain of length $T$.
Its lowest nonzero eigenvalue is $2\lambda(1-\cos(\pi/T))$ \citep{de_abreu_old_2007}, which vanishes as $O(1/T^2)$ for large $T$.
This leads to the ill-conditioned Hessian and to the small gradient in a Goldstone mode.
A more general model may couple tasks~$t$ along a sparse graph other than a chain.
The second lowest eigenvalue of the Laplacian matrix $\bL$ of a graph is called `algebraic connectivity'~\citep{de_abreu_old_2007}, and it is small in sparse graphs.


\section{Goldstone Gradient Descent}
\label{sec:method}

We now propose our solution to the slow convergence problem of Section~\ref{sec:spontsymbr}.
Algorithm~\ref{alg:gauge-gd} summarizes our Goldstone Gradient Descent (Goldstone-GD) algorithm.
We discuss details in Section~\ref{sec:algo-details}, and hyperparameters in Section~\ref{sec:algo-hyperparameters}.

The algorithm minimizes a loss function $\Ell$ of the form of Eqs.~\ref{eq:multitask-loss-1}-\ref{eq:prior_abstract}.
We alternate between standard GD (lines~\hbox{\ref{ln:begin-fullspace}-\ref{ln:standard-gradstep}} in Algorithm~\ref{alg:gauge-gd}), and a specialized minimization in the subspace of symmetry transformations (lines~\mbox{\ref{ln:begin-symspace}-\ref{ln:update-gamma}}).
The latter efficiently minimizes the symmetry breaking regularizer $\psi$ without degrading the symmetry invariant local loss functions $\ell$.
We always perform several updates of each type in a row (hyperparameters $k_1$ and $k_2$) because switching between them incurs some overhead (lines~\ref{ln:prepare-gauge} and~\ref{ln:apply-gauge}).
Algorithm~\ref{alg:gauge-gd} presents Goldstone-GD in its simplest form.
It is straight-forward to combine it with adaptive learning rates and minibatch sampling, see experiments in Section~\ref{sec:experiments}.

\begin{algorithm2e}[t]
\setlength{\hsize}{\columnwidth}
\DontPrintSemicolon
\SetKwComment{Comment}{$\triangleright\;$}{}
\SetKwFor{Repeat}{repeat}{times}{end}
\SetKwRepeat{Repeatuntil}{repeat}{until}
  \begin{tabularx}{\hsize}{@{}l@{$\,\,$}>{\raggedright\arraybackslash}X@{}}
    \textbf{Input:} & Loss function $\Ell$ of the form of Eqs.~\ref{eq:multitask-loss-1}-\ref{eq:prior_abstract}; $\qquad$
      learning rate $\rho$;
      integer hyperparameters $k_1$ and $k_2$ \\
    \textbf{Output:} & Local minimum of $\Ell$.
  \end{tabularx}\vspace{1pt}
  \nl Initialize model parameters $\bZ$ randomly\;\vspace{-1pt}
  \nl Initialize gauge fields $\tilde\bg \gets \mathbf0$\;
  \nl\Repeatuntil{\rm convergence}{
    \nl\label{ln:begin-fullspace}\Repeat{$k_1$}{
      \nl Set $\bZ \gets \bZ - \rho \nabla_{\bZ}\mathcal L(\bZ)$\; \label{ln:standard-gradstep}
      \Comment*[f]{\rm \textit{gradient step in full parameter space}}
    } \vspace{2pt}
    \nl Obtain $\mathbf M$ and $\rho'$ from Eqs.~\ref{eq:mmatrix} and \ref{eq:est-lr}\; \label{ln:prepare-gauge}
    \Comment*[f]{\rm \textit{transformation to symmetry subspace}}\\
    \nl\label{ln:begin-symspace}\Repeat{$k_2$}{
      \nl\label{ln:update-gamma}Set $\tilde\bg \gets \tilde\bg - \rho' {\bL}^{\!+}\, \nabla_{\!\tilde\bg} \Ell''(\tilde\bg; \mathbf M)$\;
      \Comment*[f]{\rm \textit{natural gradient step in symmetry subspace}}
    }
    \nl Set $Z_{t} \gets Z_{t} + (\tilde\Gamma_{\!t} - \tilde\Gamma_{\!t}^\top) Z_{t}\qquad \forall t\in\{1,\ldots,T\}$\; \label{ln:apply-gauge}
    \Comment*[f]{\rm \textit{transformation back to full parameter space}}
  }
  \caption{\hbox{Goldstone$\,$Gradient$\,$Descent$\,$(Goldstone-GD)}}
  \label{alg:gauge-gd}
\end{algorithm2e}

\subsection{Optimization in the Symmetry Subspace}
\label{sec:algo-details}

We explain lines \ref{ln:prepare-gauge}-\ref{ln:apply-gauge} of Algorithm~\ref{alg:gauge-gd}.
These steps minimize the total loss function $\Ell(\bZ)$ while restricting updates of the model parameters $\bZ$ to symmetry transformations.
Let $\mathbf R\equiv\{R_t\}_{t=1:T}$ denote $T$ orthogonal matrices.
The task is to minimize the following auxiliary loss function over $\bR$,
\begin{align} \label{eq:def-loss-prime}
    \Ell'(\bZ; \bR) \equiv \Ell(R_1 Z_1, \ldots,\! R_T Z_T) \!-\! \Ell(Z_1, \ldots,\! Z_T)
\end{align}
with the nonlinear constraint $R_t^\top\!R_t=I\;\forall t$.
If ${\bR}^*$ minimizes $\Ell'$, then updating $Z_t \gets R_t^* Z_t$ decreases the loss $\Ell$ by eliminating all Goldstone modes.
The second term on the right-hand side of Eq.~\ref{eq:def-loss-prime} does not influence the minimization as it is independent of $\bR$.
Subtracting this term makes $\Ell'$ independent of the local loss functions $\ell$:
by using Eqs.~\ref{eq:multitask-loss-1}-\ref{eq:symmetry}, we can write $\Ell'$ in terms of only the regularizer $\psi$,
\begin{align} \label{eq:loss-prime-psi}
    \Ell'(\bZ; \bR) = \psi(R_1 Z_1, \ldots,\! R_T Z_T) \!-\! \psi(Z_1, \ldots,\! Z_T).
\end{align}

\paragraph{Artificial Gauge Fields.}
We turn the constrained minimization of $\Ell'$ over $\bR$ into an unconstrained minimization using a result from the theory of Lie groups \citep{hall2015lie}.
Every special orthogonal matrix $R_t \in SO(d)$ is the matrix exponential of a skew symmetric $d\times d$ matrix $\Gamma_{\!t}$.
Here, skew symmetry means that $\Gamma_{\!t}^\top = -\Gamma_{\!t}$, and the matrix exponential function $\exp(\cdot)$ is defined by its series expansion,
\begin{align}
   \label{eq:matrix-exp}
   R_t &= \exp(\Gamma_{\!t}) \equiv I + \Gamma_{\!t} + \frac{1}{2!} \Gamma_{\!t}^2 + \frac{1}{3!} \Gamma_{\!t}^3 + \ldots
\end{align}
which is not to be confused with the componentwise exponential of $\Gamma_{\!t}$ (the term $\Gamma_{\!t}^2$ in Eq.~\ref{eq:matrix-exp} is the matrix product of $\Gamma_{\!t}$ with itself, not the componentwise square).
Eq.~\ref{eq:matrix-exp} follows from the Lie group--Lie algebra correspondence for the Lie group $SO(d)$ \citep{hall2015lie}.
Note that $R_t$ is close to the identity~$I$ if the entries of $\Gamma_{\!t}$ are small.
To enforce skew symmetry of $\Gamma_{\!t}$, we parameterize it via the skew symmetric part of an unconstrained $d \times d$ matrix $\tilde \Gamma_{\!t}$, i.e.,
\begin{align}   \label{eq:gamma-skew}
    \Gamma_{\!t} = \tilde\Gamma_{\!t} - \tilde \Gamma_{\!t}^\top.
\end{align}
We call the components of $\tilde\bg\equiv\{\tilde\Gamma_{\!t}\}_{t=1:T}$ the gauge fields, invoking an analogy to gauge theory in physics.

\paragraph{Taylor Expansion in the Gauge Fields.}
Eqs.~\ref{eq:matrix-exp}-\ref{eq:gamma-skew} parameterize a valid rotation matrix $R_t$ in terms of an arbitrary $d\times d$ matrix $\tilde\Gamma_{\!t}$.
This turns the constrained minimization of $\Ell'$ into an unconstrained one.
However, the matrix-exponential function in Eq.~\ref{eq:matrix-exp} is numerically expensive, and its derivative is complicated because the group $SO(d)$ is non-abelian.
We simplify the problem by introducing an approximateion.
As the model parameters $\bZ$ approach the minimum of $\Ell$, the optimal rotations $\bR^*$ that minimize $\Ell'$ converge to the identity, and thus the gauge fields converge to zero.
In this limit, the approximation becomes exact.


We approximate the auxiliary loss function $\Ell'$ by a second order Taylor expansion $\Ell''$.
In detail, we truncate Eq.~\ref{eq:matrix-exp} after the term quadratic in $\Gamma_{\!t}$ and insert the truncated series into Eq.~\ref{eq:loss-prime-psi}.
We multiply out the quadratic form in the prior $\psi$, Eq.~\ref{eq:prior_abstract}, and neglect again all terms of higher than quadratic order in $\bg$.
Using the skew symmetry of $\Gamma_{\!t}$ and the symmetry of the Laplacian matrix $\bL = \mathbf D-\mathbf A$, we find
\begin{align} \label{eq:def-ell2prime}
  \Ell''(\tilde\bg;\mathbf M)
  \!&=\! \sum_{t,t'} A_{tt'} {\rm Tr}\!\left[
      \left( \!\Gamma_{\!t'} \!+\! \frac12 (\Gamma_{\!t'} \!-\! \Gamma_{\!t}) \Gamma_{\!t} \!\right) \!M_{tt'} \!\right]
\end{align}
where the trace runs over the embedding space, and for each $t,t'\in\{1,\ldots, T\}$, we define the matrix $M_{tt'} \in \IR^{d\times d}$,
\begin{align} \label{eq:mmatrix}
  M_{tt'} &\equiv \sum_{i=1}^N z_{t,i}z_{t',i}^\top.
\end{align}
We evaluate the matrices $M_{tt'}$ on line~\ref{ln:prepare-gauge} in Algorithm~\ref{alg:gauge-gd}.
Note that the adjacency matrix $\mathbf A$ is sparse, and that we only need to obtain those matrices $M_{tt'}$ for which $A_{tt'}\neq0$.

We describe the numerical minimization of $\Ell''$ below.
Once we obtain gauge fields $\tilde{\bg}^*$ that minimize $\Ell''$, the optimal update step for the model parameters would be $Z_t \gets \exp(\tilde\Gamma^*_{\!t}-\tilde\Gamma^{*\top}_{\!t}) Z_t$.
For efficiency, we truncate the matrix exponential function $\exp(\cdot)$ after the linear term, resulting in line~\ref{ln:apply-gauge} of Algorithm~\ref{alg:gauge-gd}.
We do not reset the gauge fields $\tilde\bg$ to zero after updating $\bZ$, so that the next minimization of $\Ell''$ starts with preinitialized $\tilde\bg$.
This turned out to speed up convergence in our experiments, possibly because $\tilde\bg$ acts like a momentum in the symmetry subspace.

\paragraph{Natural Gradients.}
Lines~\ref{ln:begin-symspace}-\ref{ln:update-gamma} in Algorithm~\ref{alg:gauge-gd} minimize $\Ell''$ over the gauge fields $\tilde\bg$ using GD.
We speed up convergence using the fact that $\Ell''$ depends only on the prior $\psi$ and not on $\ell$.
Since we know the Hessian of $\psi$, we can use natural gradients \citep{amari1998natural}, resulting in the update step
\begin{align} \label{eq:nat-grad}
  \tilde\bg \gets \tilde\bg - \rho' \bL^{\!+}\, \nabla_{\!\tilde\bg} \Ell''(\tilde\bg; \mathbf M)
\end{align}
where $\rho'$ is a constant learning rate and $\bL^{\!+}$ is the pseudoinverse of the Laplacian matrix $\bL$.
We obtain $\bL^{\!+}$ by taking the eigendecomposition of $\bL$ and inverting the eigenvalues, except for the single zero eigenvalue corresponding to (irrelevant) global rotations, which we leave at zero.
$\bL^{\!+}$ has to be obtained only once before entering the training loop.

\paragraph{Learning Rate.}
We find that we can automatically set $\rho'$ in Eq.~\ref{eq:nat-grad} to a value that leads to fast convergence,
\begin{align} \label{eq:est-lr}
  \rho' = \frac{1}{TN \langle Z^2\rangle}
  \quad\text{with}\quad
  \langle Z^2\rangle \equiv \frac{1}{TNd} \sum_{t,i} {||z_{t,i}||}_2^2.
\end{align}
We arrive at this choice of learning rate by estimating the Hessian of $\Ell''$.
The preconditioning with $\bL^{\!+}$ in Eq.~\ref{eq:nat-grad} takes into account the structure of the Hessian in $t$-space, which enters $\Ell''$ in Eq.~\ref{eq:def-ell2prime} via the adjacency matrix $\mathbf A$.
The remaining factor $M_{tt'}$, defined in Eq.~\ref{eq:mmatrix}, is quadratic in the components of $\bZ$ and linear in $N$.
This suggests a learning rate $\rho' \propto 1/(N\langle Z^2\rangle)$.
We find empirically for large models that the $t$-dependency of $M_{tt'}$ leads to a small mismatch between $\bL$ and the Hessian of $\Ell''$.
The more conservative choice of learning rate in Eq.~\ref{eq:est-lr} leads to fast convergence of the gauge fields in all our experiments.

\subsection{Hyperparameters}
\label{sec:algo-hyperparameters}

\begin{table}[t]
\vskip -0.085in
\caption{Runtimes of operations in Goldstone-GD ({\sc L}$=$line in Algorithm~\ref{alg:gauge-gd}; \#$=$frequency of execution; $T$=no.~of time steps; $N$=input dimension; $d$=embedding dimension; $k_1,k_2$=hyperparameters).}
\label{table:complexities}
\vskip 0.15in
\begin{center}
\begin{small}
\begin{sc}
\begin{tabularx}{\columnwidth}{@{}X@{}r@{\ \ }l@{\ \ \ }l@{\ \ }l@{}X@{}}
\toprule
&L & Operation & Complexity & \ \ \# & \\
\midrule
&\ref{ln:standard-gradstep} & \rm gradient step in full param.~space & \rm model dependent & $\times k_1$ & \\
&\ref{ln:prepare-gauge} & \rm transformation to symmetry space & $O(TNd^2)$ & $\times 1$ & \\
&\ref{ln:update-gamma} & \rm nat.~grad.~step in symmetry space & $O(Td^3 + T^2d^2)$ & $\times k_2$ & \\
&\ref{ln:apply-gauge} & \rm transformation to full param.~space & $O(TNd^2)$ & $\times 1$ & \\
\bottomrule
\end{tabularx}
\end{sc}
\end{small}
\end{center}
\vskip -0.1in
\end{table}

Goldstone-GD has two integer hyperparameters, $k_1$ and $k_2$, which control the frequency of execution of each operation.
Table \ref{table:complexities} lists the computational complexity of each operation, assuming that the sparse adjacency matrix $\mathbf A$ has $O(T)$ nonzero entries, as is the case in Markovian time series models (Eq.~\ref{eq:tridiagonal-l}).
Note that the embedding dimension $d$ is typically orders of magnitude smaller than the input dimension $N$.
Therefore, update steps in the symmetry subspace (line~\ref{ln:update-gamma}) are cheap.
In our experiments, we always set $k_1$ and $k_2$ such that the
overhead from lines~\ref{ln:prepare-gauge}-\ref{ln:apply-gauge} is less than $10\%$.


\section{Experiments}
\label{sec:experiments}

We evaluate the proposed Goldstone-GD optimization algorithm on the three example models introduced in Section~\ref{sec:models}.
We compare Goldstone-GD to standard GD, to AdaGrad \citep{duchi_adaptive_2011}, and to Adam \citep{kingma_adam_2014}.
Goldstone-GD converges orders of magnitude faster and fits more interpretable word embeddings.

\subsection{Visualizing Goldstone Modes With Artificial Data}
\label{sec:exp-artificial-data}

\paragraph{Model and Data Preparation.}
We fit the dynamic Gaussian matrix factorization model defined in Eqs.~\ref{eq:matfact-z}-\ref{eq:time-series-prior} in Section~\ref{sec:models} to small scale artificial data.
In order to visualize Goldstone modes in the embedding space we choose an embedding dimension of $d=3$ and, for this experiment only, we fit the model to time independent-data.
This allows us to monitor convergence since we know that the matrices $U_t^*$ and $V_t^*$ that minimize the loss are also time-independent.
We generate artificial data for the matrix $X\in\mathbb R^{10\times 10}$ by drawing the components of two matrices $\bar U,\bar V\in\IR^{3\times 10}$ from a standard normal distribution, forming $\bar U^\top \bar V$, and adding uncorrelated Gaussian noise with variance $10^{-3}$.

\paragraph{Hyperparameters.}
We use $T=30$ time steps and a coupling strength of $\lambda=10$.
We train the model with standard GD (baseline) and with Goldstone-GD with $k_1=50$ and $k_2=10$.
We find fastest convergence for the baseline method if we clip the gradients to an interval $[-\bar g,\bar g]$ and use a decreasing learning rate $\rho_s = \rho_0 (\bar s/(s+\bar s))^{0.7}$ despite the noise-free gradient.
Here, $s$ is the training iteration.
We optimize the hyperparameters for fastest convergence in the baseline and find $\bar g=0.01$, $\rho_0=1$, and $\bar s=100$.

\paragraph{Results.}
Figure~\ref{fig:matrixfact-mini} compares convergence in the two algorithms.
We discussed the figure at the end of Section~\ref{sec:spontsymbr}.
In summary, Goldstone-GD converges an order of magnitude faster even in this small scale setup that allows only for three different kinds of Goldstone modes (the skew symmetric gauge fields $\Gamma_{\!t}$ have only $d(d-1)/2=3$ independent parameters).
Once the minimization finds minima of the local losses $\ell$, differences in the total loss $\Ell$ between the two algorithms are small since Goldstone modes contribute only little to $\Ell$ (this is why they decay slowly in GD).

\subsection{MovieLens Recommendations}
\label{sec:exp-movielens}

\paragraph{Model and Data Set.}
We fit the sparse dynamic Bernoulli factorization model defined in Eqs.~\ref{eq:time-series-prior}-\ref{eq:sparse-matfact-localloss} in Section~\ref{sec:models} to the Movielens 20M data set\footnote{\url{https://grouplens.org/datasets/movielens/20m/}} \citep{harper2016movielens}.
We use embedding dimension $d=30$, coupling strength $\lambda=10$, and regularizer $\gamma=1$.
The data set consists of $20$ million reviews of $27{,}000$ movies by $138{,}000$ users with time stamps from 1995 to 2015.
We binarize the ratings by splitting at the median, discarding ratings at the median, and we slice the remaining $18$ million data points into $T=100$ time bins of equal duration.
We split randomly across all bins into $50\%$ training, $20\%$ validation, and $30\%$ test set.

\begin{figure}[t]
\vskip 0.04in
\begin{center}
\centerline{\includegraphics[width=\columnwidth]{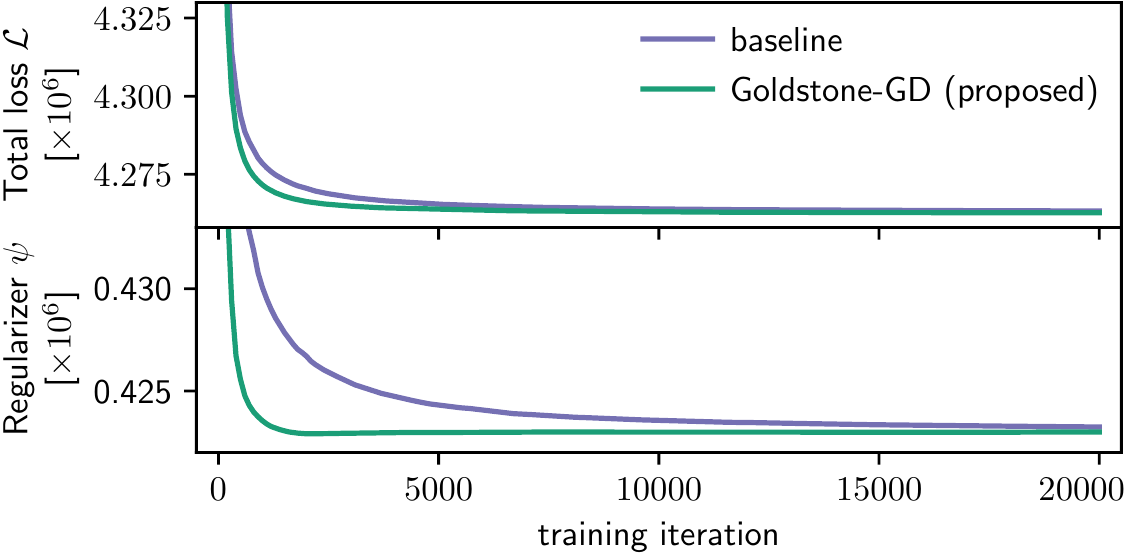}}
\caption{Training curves for MovieLens recommendations (sparse dynamic matrix factorization; Section~\ref{sec:exp-movielens}).
Three different random initializations lead to indistinguishable results at this scale.
$x$-axis not accounting for a $1\%$ runtime overhead in Goldstone-GD.}
\label{fig:movielens-learningcurves}
\end{center}
\vskip -0.2in
\end{figure}

\paragraph{Baseline and Hyperparameters.}
We compare the proposed Goldstone-GD algorithm to GD with AdaGrad \citep{duchi_adaptive_2011} with a learning rate prefactor of $1$ obtained from cross-validation.
Similar to Goldstone-GD, AdaGrad is designed to escape shallow valleys of the loss, but it uses only diagonal preconditioning.
We compare to Goldstone-GD with $k_1=100$ and $k_2=10$, using the same AdaGrad optimizer for update steps in the full parameter space.

\begin{table*}[t]
\vskip -0.085in
\caption{Word aging: We translate modern words to the year $1800$ using the shared representation space of Dynamic Word Embeddings.}
\label{table:word-aging}
\vskip 0.15in
\begin{center}
\begin{small}
\begin{sc}
\begin{tabularx}{\textwidth}{l@{}X@{}l@{}X@{}l@{}X@{}}
\toprule
Query && Goldstone-GD && Baseline & \\
\midrule
\rm car && \rm boat, saddle, canoe, wagon, box && \rm shell, roof, ceiling, choir, central &\\
\rm computer && \rm perspective, telescope, needle, mathematical, camera && \rm organism, disturbing, sexual, rendering, bad &\\
\rm DNA && \rm potassium, chemical, sodium, molecules, displacement && \rm operates, differs, sharing, takes, keeps  &\\
\rm electricity	&& \rm vapor, virus, friction, fluid, molecular && \rm exercising, inherent, seeks, takes, protect &\\
\rm tuberculosis && \rm chronic, paralysis, irritation, disease, vomiting && \rm trained, uniformly, extinguished, emerged, widely & \\
\bottomrule
\end{tabularx}
\end{sc}
\end{small}
\end{center}
\vskip -0.1in
\end{table*}

\paragraph{Results.}
The additional operations in Goldstone-GD lead to a $1\%$ overhead in runtime.
The upper panel in Figure~\ref{fig:movielens-learningcurves} shows training curves for the loss $\Ell$ using the baseline (purple) and Goldstone-GD (green).
The loss $\Ell$ drops faster in Goldstone-GD, but differences in terms of the full loss $\Ell$ are small because the local loss functions $\ell$ are much larger than the regularizer $\psi$ in this experiment.
The lower panel of Figure~\ref{fig:movielens-learningcurves} shows only $\psi$.
Both algorithms converge to the same value of~$\psi$, but Goldstone-GD converges at least an order of magnitude faster.
The difference in value is small because Goldstone modes contribute little to $\psi$.
They can, however, have a large influence on the parameter values, as we show next in experiments with Dynamic Word Embeddings.

\subsection{Dynamic Word Embeddings}
\label{sec:exp-dwe}

\paragraph{Model and Data Set.}
We perform variational inference \citep{ranganath2014black} in Dynamic Word Embeddings (DWE), see Section~\ref{sec:models}.
We fit the model to digitized books from the years $1800$ to $2008$ in the Google Books corpus\footnote{\url{http://storage.googleapis.com/books/ngrams/books/datasetsv2.html}} \citep{michel_quantitative_2011} (approximately $10^{10}$ words).
We follow \citep{bamler_dynamic_2017} for data preparation, resulting in a vocabulary size of $10{,}000$, a training set of $T=188$ time step, and a test set of $21$ time steps.
The DWE paper proposes two inference algorithms: filtering and smoothing.
We use the smoothing algorithm, which has better predictive performance than filtering but suffers from Goldstone modes.
We set the embedding dimension to $d=100$ due to hardware constraints and train for $10{,}000$ iterations using an Adam optimizer~\citep{kingma_adam_2014} with a decaying prefactor of the adaptive learning rate, $\rho_s=\rho_0 (\bar s/(s + \bar s))^{0.7}$, where $s$ is the training iteration, $\rho_0=0.1$, and $\bar s=1000$.
We find that this leads to better convergence than a constant prefactor.
All other hyperparameters are the same as in~\citep{bamler_dynamic_2017}.
We compare the baseline to Goldstone-GD using the same learning rate schedule and $k_1=k_2=10$, which leads to an $8\%$ runtime overhead.

\paragraph{Results.}
By eliminating Goldstone modes, Goldstone-GD makes word embeddings comparable across the time dimension of the model.
We demonstrate this in Table~\ref{table:word-aging}, which shows the result of `aging' modern words, i.e., translating them from modern English to the English language of the year $1800$.
For each query word $i$, we report the five words $i'$ whose embedding vectors $u_{i'\!,1}$ at the first time step (year $1800$) have largest overlap with the embedding vector $u_{i,T}$ of the query word at the last time step (year $2008$).
Overlap is measured in cosine distance (normalized scalar product), between the means of $u_{i,T}$ and $u_{i'\!,1}$ under the variational distribution.

\begin{figure}[t]
\vskip 0.04in
\begin{center}
\centerline{\includegraphics[width=\columnwidth]{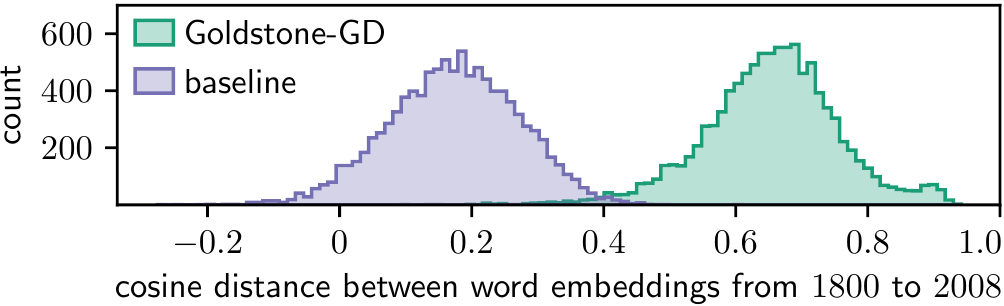}}
\caption{Cosine distance between word embeddings from the first and last year of the training data in Dynamic Word Embeddings.}
\label{fig:dwe-histogram}
\end{center}
\vskip -0.2in
\end{figure}

Goldstone-GD finds words that are plausible for the year $1800$ while still being related to the query (e.g., means of transportation in a query for `car').
By contrast, the baseline method fails to find plausible results.
Figure~\ref{fig:dwe-histogram} provides more insight into the failure of the baseline method.
It shows histograms of the cosine distance between word embeddings $u_{i,1}$ and $u_{i,T}$ for the same word $i$ from the first to the last time step.
In Goldstone-GD (green), most embeddings have a large overlap because the meaning of most words does not change drastically over time.
By contrast, in the baseline (purple), no embeddings overlap by more than $60\%$ between $1800$ and $2008$, and some embeddings even change their orientation (negative overlap).
We explain this counterintuitive result with the presence of Goldstone modes, i.e., the entire embedding spaces are rotated against each other.

For a quantitative comparison, we evaluate the predictive log-likelihood of the test set under the posterior mean, and find slightly better predictive performance with Goldstone-GD ($-0.5317$ vs.~$-0.5323$ per test point).
The improvement is small because the training set is so large that the influence of the symmetry breaking regularizer is dwarfed in all but the symmetry directions by the log-likelihood of the data.
The main advantage of Goldstone-GD are the more interpretable embeddings, as demonstrated in Table~\ref{table:word-aging}.


\section{Conclusions}
\label{sec:conclusions}

We identified a slow convergence problem in representation learning models with a continuous symmetry and a Markovian time series prior, and we solved the problem with a new optimization algorithm, Goldstone-GD.
The algorithm separates the minimization in the symmetry subspace from the remaining coordinate directions.
Our experiments showed that Goldstone-GD converges orders of magnitude faster and fits more interpretable embedding vectors, which can be compared across the time dimension of a model.
Since continuous symmetries are common in representation learning, we believe that gauge theories and, more broadly, the theory of Lie groups are more widely useful in machine learning.


\section*{Acknowledgements}

We thank Ari Pakman for valuable and detailed feedback that greatly improved the manuscript.

\bibliography{references}

\begin{thebibliography}{36}
\providecommand{\natexlab}[1]{#1}
\providecommand{\url}[1]{\texttt{#1}}
\expandafter\ifx\csname urlstyle\endcsname\relax
  \providecommand{\doi}[1]{doi: #1}\else
  \providecommand{\doi}{doi: \begingroup \urlstyle{rm}\Url}\fi

\bibitem[Altland \& Simons(2010)Altland and Simons]{altland2010condensed}
Altland, A. and Simons, B.~D.
\newblock \emph{Condensed matter field theory}.
\newblock Cambridge University Press, 2010.

\bibitem[Amari(1998)]{amari1998natural}
Amari, S.-I.
\newblock Natural gradient works efficiently in learning.
\newblock \emph{Neural computation}, 10\penalty0 (2):\penalty0 251--276, 1998.

\bibitem[Badrinarayanan et~al.(2015)Badrinarayanan, Mishra, and
  Cipolla]{badrinarayanan2015understanding}
Badrinarayanan, V., Mishra, B., and Cipolla, R.
\newblock Understanding symmetries in deep networks.
\newblock \emph{arXiv preprint arXiv:1511.01029}, 2015.

\bibitem[Bamler \& Mandt(2017)Bamler and Mandt]{bamler_dynamic_2017}
Bamler, R. and Mandt, S.
\newblock Dynamic word embeddings.
\newblock In \emph{Proceedings of the 34th International Conference on Machine
  Learning (ICML)}, pp.\  380--389, 2017.

\bibitem[Barkan(2017)]{barkan2016bayesian}
Barkan, O.
\newblock Bayesian {Neural} {Word} {Embedding}.
\newblock In \emph{Proceedings of the Thirty-First AAAI Conference on
  Artificial Intelligence}, 2017.

\bibitem[Bui et~al.(2013)Bui, Huynh, and Riedel]{bui2012automorphism}
Bui, H.~H., Huynh, T.~N., and Riedel, S.
\newblock Automorphism groups of graphical models and lifted variational
  inference.
\newblock In \emph{Proceedings of the Twenty-Ninth Conference on Uncertainty in
  Artificial Intelligence}, pp.\  132--141, 2013.

\bibitem[Charlin et~al.(2015)Charlin, Ranganath, McInerney, and
  Blei]{charlin2015dynamic}
Charlin, L., Ranganath, R., McInerney, J., and Blei, D.~M.
\newblock Dynamic {Poisson} factorization.
\newblock In \emph{Proceedings of the 9th ACM Conference on Recommender
  Systems}, pp.\  155--162, 2015.

\bibitem[Choi et~al.(1999)Choi, Amari, Cichocki, and Liu]{choi1999natural}
Choi, S., Amari, S., Cichocki, A., and Liu, R.-w.
\newblock Natural gradient learning with a nonholonomic constraint for blind
  deconvolution of multiple channels.
\newblock In \emph{First International Workshop on Independent Component
  Analysis and Signal Separation}, pp.\  371--376, 1999.

\bibitem[de~Abreu(2007)]{de_abreu_old_2007}
de~Abreu, N. M.~M.
\newblock Old and new results on algebraic connectivity of graphs.
\newblock \emph{Linear Algebra and its Applications}, 423\penalty0
  (1):\penalty0 53--73, 2007.

\bibitem[Duchi et~al.(2011)Duchi, Hazan, and Singer]{duchi_adaptive_2011}
Duchi, J., Hazan, E., and Singer, Y.
\newblock Adaptive {Subgradient} {Methods} for {Online} {Learning} and
  {Stochastic} {Optimization}.
\newblock \emph{Journal of Machine Learning Research}, 12:\penalty0 2121--2159,
  2011.

\bibitem[Gens \& Domingos(2014)Gens and Domingos]{gens2014deep}
Gens, R. and Domingos, P.~M.
\newblock Deep symmetry networks.
\newblock In \emph{Advances in Neural Information Processing Systems 27}, pp.\
  2537--2545. 2014.

\bibitem[Gopalan et~al.(2015)Gopalan, Hofman, and Blei]{gopalan2015scalable}
Gopalan, P., Hofman, J.~M., and Blei, D.~M.
\newblock Scalable recommendation with hierarchical {Poisson} factorization.
\newblock In \emph{UAI}, pp.\  326--335, 2015.

\bibitem[Hall(2015)]{hall2015lie}
Hall, B.
\newblock \emph{Lie groups, Lie algebras, and representations: an elementary
  introduction}, volume 222.
\newblock Springer, 2015.

\bibitem[Harper \& Konstan(2016)Harper and Konstan]{harper2016movielens}
Harper, F.~M. and Konstan, J.~A.
\newblock The {MovieLens} {Datasets}: {History} and {Context}.
\newblock \emph{ACM Transactions on Interactive Intelligent Systems (TiiS)},
  5\penalty0 (4):\penalty0 19, 2016.

\bibitem[Hoffman et~al.(2013)Hoffman, Blei, Wang, and
  Paisley]{hoffman_stochastic_2013}
Hoffman, M.~D., Blei, D.~M., Wang, C., and Paisley, J.~W.
\newblock Stochastic {Variational} {Inference}.
\newblock \emph{Journal of Machine Learning Research}, 14\penalty0
  (1):\penalty0 1303--1347, 2013.

\bibitem[Kingma \& Ba(2014)Kingma and Ba]{kingma_adam_2014}
Kingma, D. and Ba, J.
\newblock Adam: {A} {Method} for {Stochastic} {Optimization}.
\newblock In \emph{Proceedings of the 3rd International Conference for Learning
  Representations (ICLR)}, 2014.

\bibitem[Koren(2010)]{koren2010collaborative}
Koren, Y.
\newblock Collaborative filtering with temporal dynamics.
\newblock \emph{Communications of the ACM}, 53\penalty0 (4):\penalty0 89--97,
  2010.

\bibitem[LeCun et~al.(1998)LeCun, Bottou, Bengio, and Haffner]{Lecun98gradient}
LeCun, Y., Bottou, L., Bengio, Y., and Haffner, P.
\newblock Gradient-based learning applied to document recognition.
\newblock \emph{Proceedings of the IEEE}, 86\penalty0 (11):\penalty0
  2278--2324, 1998.

\bibitem[Lu et~al.(2009)Lu, Agarwal, and Dhillon]{lu2009aspatiot}
Lu, Z., Agarwal, D., and Dhillon, I.~S.
\newblock A spatio--temporal approach to collaborative filtering.
\newblock In \emph{ACM Conference on Recommender Systems (RecSys)}, 2009.

\bibitem[Martens(2014)]{martens2014new}
Martens, J.
\newblock New insights and perspectives on the natural gradient method.
\newblock \emph{arXiv preprint arXiv:1412.1193}, 2014.

\bibitem[Michel et~al.(2011)Michel, Shen, Aiden, Veres, Gray, Pickett, Hoiberg,
  Clancy, Norvig, Orwant, et~al.]{michel_quantitative_2011}
Michel, J.-B., Shen, Y.~K., Aiden, A.~P., Veres, A., Gray, M.~K., Pickett,
  J.~P., Hoiberg, D., Clancy, D., Norvig, P., Orwant, J., et~al.
\newblock Quantitative {Analysis} of {Culture} {Using} {Millions} of
  {Digitized} {Books}.
\newblock \emph{Science}, 331\penalty0 (6014):\penalty0 176--182, 2011.

\bibitem[Mikolov et~al.(2013{\natexlab{a}})Mikolov, Chen, Corrado, and
  Dean]{mikolov_efficient_2013}
Mikolov, T., Chen, K., Corrado, G., and Dean, J.
\newblock Efficient {Estimation} of {Word} {Representations} in {Vector}
  {Space}.
\newblock \emph{arXiv preprint arXiv:1301.3781}, 2013{\natexlab{a}}.

\bibitem[Mikolov et~al.(2013{\natexlab{b}})Mikolov, Sutskever, Chen, Corrado,
  and Dean]{mikolov_distributed_2013}
Mikolov, T., Sutskever, I., Chen, K., Corrado, G.~S., and Dean, J.
\newblock {Distributed} {Representations} of {Words} and {Phrases} and their
  {Compositionality}.
\newblock In \emph{Advances in Neural Information Processing Systems 26}, pp.\
  3111--3119. 2013{\natexlab{b}}.

\bibitem[Mnih \& Salakhutdinov(2008)Mnih and
  Salakhutdinov]{mnih2008probabilistic}
Mnih, A. and Salakhutdinov, R.~R.
\newblock Probabilistic matrix factorization.
\newblock In \emph{Advances in neural information processing systems}, pp.\
  1257--1264, 2008.

\bibitem[Murphy(2012)]{murphy2012machine}
Murphy, K.~P.
\newblock \emph{Machine {Learning}: {A} {Probabilistic} {Perspective}}.
\newblock MIT Press, 2012.

\bibitem[Noessner et~al.(2013)Noessner, Niepert, and
  Stuckenschmidt]{noessner2013rockit}
Noessner, J., Niepert, M., and Stuckenschmidt, H.
\newblock Rockit: Exploiting parallelism and symmetry for map inference in
  statistical relational models.
\newblock In \emph{AAAI Workshop: Statistical Relational Artificial
  Intelligence}, 2013.

\bibitem[Pascanu \& Bengio(2013)Pascanu and Bengio]{pascanu2013revisiting}
Pascanu, R. and Bengio, Y.
\newblock Revisiting natural gradient for deep networks.
\newblock \emph{arXiv preprint arXiv:1301.3584}, 2013.

\bibitem[Pennington et~al.(2014)Pennington, Socher, and
  Manning]{pennington2014glove}
Pennington, J., Socher, R., and Manning, C.
\newblock Glove: Global vectors for word representation.
\newblock In \emph{Proceedings of the 2014 conference on empirical methods in
  natural language processing (EMNLP)}, pp.\  1532--1543, 2014.

\bibitem[Poignard et~al.(2018)Poignard, Pereira, and Pade]{poignard2018spectra}
Poignard, C., Pereira, T., and Pade, J.~P.
\newblock Spectra of laplacian matrices of weighted graphs: structural
  genericity properties.
\newblock \emph{SIAM Journal on Applied Mathematics}, 78\penalty0 (1):\penalty0
  372--394, 2018.

\bibitem[Ranganath et~al.(2014)Ranganath, Gerrish, and
  Blei]{ranganath2014black}
Ranganath, R., Gerrish, S., and Blei, D.
\newblock Black box variational inference.
\newblock In \emph{Artificial Intelligence and Statistics}, pp.\  814--822,
  2014.

\bibitem[Rudolph \& Blei(2018)Rudolph and Blei]{rudolph2017dynamic}
Rudolph, M. and Blei, D.
\newblock Dynamic embeddings for language evolution.
\newblock In \emph{Proceedings of the 2018 World Wide Web Conference on World
  Wide Web}, pp.\  1003--1011, 2018.

\bibitem[Sun et~al.(2016)Sun, Guo, Lan, Xu, and Cheng]{sun2016sparse}
Sun, F., Guo, J., Lan, Y., Xu, J., and Cheng, X.
\newblock Sparse word embeddings using $l_1$ regularized online learning.
\newblock In \emph{Proceedings of the Twenty-Fifth International Joint
  Conference on Artificial Intelligence}, pp.\  2915--2921, 2016.

\bibitem[Sun et~al.(2012)Sun, Varshney, and Subbian]{sun2012dynamic}
Sun, J.~Z., Varshney, K.~R., and Subbian, K.
\newblock Dynamic matrix factorization: A state space approach.
\newblock In \emph{2012 IEEE International Conference on Acoustics, Speech and
  Signal Processing (ICASSP)}, pp.\  1897--1900, 2012.

\bibitem[Vinyals \& Povey(2012)Vinyals and Povey]{vinyals2012krylov}
Vinyals, O. and Povey, D.
\newblock Krylov subspace descent for deep learning.
\newblock In \emph{Artificial Intelligence and Statistics}, pp.\  1261--1268,
  2012.

\bibitem[Zaheer et~al.(2017)Zaheer, Kottur, Ravanbakhsh, Poczos, Salakhutdinov,
  and Smola]{zaheer2017deep}
Zaheer, M., Kottur, S., Ravanbakhsh, S., Poczos, B., Salakhutdinov, R.~R., and
  Smola, A.~J.
\newblock Deep sets.
\newblock In \emph{Advances in Neural Information Processing Systems}, pp.\
  3394--3404, 2017.

\bibitem[Zeiler(2012)]{zeiler2012adadelta}
Zeiler, M.~D.
\newblock {ADADELTA}: an {Adaptive} {Learning} {Rate} {Method}.
\newblock \emph{arXiv preprint arXiv:1212.5701}, 2012.

\end{thebibliography}
\bibliographystyle{icml2018}

\end{document}